\documentclass[10pt, a4paper]{article}

\usepackage{booktabs}
\usepackage{hyperref}
\usepackage{tabularx}
\usepackage{paralist}
\usepackage{graphicx}
\usepackage{pgffor}
\usepackage{xcolor}
\usepackage{xspace}
\usepackage{multirow}

\newcommand{\corpushi}[1]{\textsc{#1}}
\newcommand{\contarga}{\corpushi{ContArgA}\xspace}
\newcommand{\caaf}{\corpushi{CAAF}\xspace}
\newcommand{\emohi}[1]{\textit{#1}\xspace}

\newcommand{\anger}{\emohi{anger}}

\newcommand{\joy}{\emohi{joy}}

\newcommand{\fear}{\emohi{fear}}
\newcommand{\Fear}{\emohi{Fear}}
\newcommand{\sadness}{\emohi{sadness}}

\newcommand{\disgust}{\emohi{disgust}}

\newcommand{\surprise}{\emohi{surprise}}
\newcommand{\Surprise}{\emohi{Surprise}}
\newcommand{\pride}{\emohi{pride}}
\newcommand{\Pride}{\emohi{Pride}}
\newcommand{\relief}{\emohi{relief}}

\newcommand{\shame}{\emohi{shame}}

\newcommand{\guilt}{\emohi{guilt}}

\newcommand{\trust}{\emohi{trust}}

\newcommand{\suddenness}{\emohi{suddenness}}

\newcommand{\suppression}{\emohi{suppression}}

\newcommand{\familiarity}{\emohi{familiarity}}

\newcommand{\pleasantness}{\emohi{pleasantness}}
\newcommand{\Pleasantness}{\emohi{Pleasantness}}
\newcommand{\unpleasantness}{\emohi{unpleasantness}}

\newcommand{\consimportnace}{\emohi{consequencial importance}}

\newcommand{\posconseq}{\emohi{positive consequentiality}}

\newcommand{\negconseq}{\emohi{negative consequentiality}}

\newcommand{\conseqmanage}{\emohi{consequence manageability}}

\newcommand{\intcheck}{\emohi{internal check}}

\newcommand{\excheck}{\emohi{external check}}

\newcommand{\responseurgency}{\emohi{response urgency}}

\newcommand{\cogeffort}{\emohi{cognitive effort}}

\newcommand{\argintcheck}{\emohi{argument internal check}}

\newcommand{\argexcheck}{\emohi{argument external check}}

\makeatletter
\renewcommand\paragraph{\@startsection{paragraph}{4}{\z@}%
  {0.8ex \@plus1ex \@minus.2ex}%
  {-1em}%
  {\normalfont\normalsize\bfseries}}
\makeatother

\usepackage[final]{lrec2026} 

\title{Trust Me, I Can Convince You: The Contextualized Argument
  Appraisal Framework and the \contarga Corpus}

\name{Lynn Greschner, Sabine Weber, Roman Klinger} 

\address{Fundamentals of Natural Language Processing, University of Bamberg, Germany\\
  \{firstname.lastname\}@uni-bamberg.de\\}

\abstract{%
  Emotions that somebody develops based on an argument do not only
  depend on the argument itself -- they are also influenced by a
  subjective evaluation of the argument's potential impact on the
  self. For instance, an argument to ban plastic bottles might cause
  fear of losing a job for a bottle industry worker, which lowers the
  convincingness -- presumably independent of its content. While
  binary emotionality of arguments has been studied, such cognitive
  appraisal models have only been proposed in other subtasks of
  emotion analysis, but not in the context of arguments and their
  convincingness. To fill this research gap, we propose the
  Contextualized Argument Appraisal Framework to model the interplay
  between the sender, receiver, and argument. We adapt established
  appraisal models from psychology to argument mining, including
  argument pleasantness, familiarity, response urgency, and expected
  effort, as well as convincingness variables. To evaluate the
  framework and pave the way for computational modeling, we develop a
  novel role-playing-based annotation setup, mimicking real-world
  exposure to arguments. Participants disclose their emotion, explain
  the main cause, the argument appraisal, and the perceived
  convincingness. To consider the subjective nature of such
  annotations, we also collect demographic data and personality traits
  of both the participants and ask them to disclose the same variables
  for their perception of the argument sender.  The analysis of the
  resulting \contarga corpus of 4000 annotations reveals that
  convincingness is positively correlated with positive emotions
  (e.g., trust) and negatively correlated with negative emotions
  (e.g., anger). The appraisal variables particularly point to the
  importance of the annotator's familiarity with the argument. \\%
  \newline %
}

\begin{document}

\maketitleabstract

\section{Introduction}
People are frequently exposed to argumentation, where the main goal is
persuasion \citep{habernal-gurevych-2016-makes}. Argument quality
plays a key role in this process, depending on logical structure
(logos), speaker credibility (ethos), emotional appeal (pathos), and
contextual relevance (kairos) \citep{aristotle1991rhetoric,
  schiappa2013argumentation}.  While there is a substantial amount of
research on the overall argument quality and convincingness
\citep{habernal-gurevych-2016-argument, gleize-etal-2019-convinced,
  wachsmuth-etal-2017-computational, lauscher-etal-2020-rhetoric}, the
emotional appeal of arguments remains underexplored in the natural
language processing (NLP) community.

\begin{figure}
  \centering
  \includegraphics[width=0.9\linewidth]{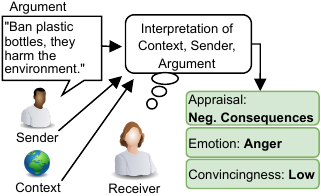} 
  \caption{Illustration of the Contextualized Argument Appraisal
    Framework. The perception of the argument, context and sender
    are cognitively evaluated and assessed as having a negative
    consequence for the receiver (e.g., losing a job
    if plastic bottles were banned).}
  \label{fig:framework}
\end{figure} 

One potential reason is that capturing emotions in text is a
challenging task due to their contextualized and implicit nature
\citep{casel-etal-2021-emotion,klinger-etal-2018-iest,koga-etal-2024-forecasting-implicit,lee-lau-2020-event}. This
is particularly true in arguments, which rarely directly express an
emotion that is intended to be caused. Figure~\ref{fig:framework}
shows an example in our framework. While the argument appears
non-emotional, various factors influence the emotional response: prior
beliefs of the receiver, their personality, the relationship with the
sender and their traits, or the content of the argument
itself. Appraisal theories serve as a suitable basis to model the
associated cognitive component \citep{ellsworth2003appraisal}. While
they received some attention in event-centered emotion analysis
\citep{klinger-2023-event}, there is neither a psychological framework
of argument appraisal, nor is there work in NLP on fine-grained
modeling of emotions in arguments.

We close this gap by proposing the Contextualized Argument Appraisal
Framework which focuses on the subjective perception of arguments. It
encompasses \textit{emotion categories}, a dedicated set of
\textit{appraisal variables} such as argument familiarity, expected
effort, or the consequences on the own goals, and the perceived
\textit{convincingness}, all in context of variables of the
(perceived) argument sender, the receiver, and the argument text
itself. The framework therefore is an emotion argument analysis
approach under the paradigm of perspectivism
\citep{frenda_perspectivist_2025}.

In this paper, we use the proposed framework and data from the
corresponding study to answer the following research questions:
\begin{compactitem}
    \item \textbf{RQ1:} How do discrete emotions and appraisals correlate with argument convincingness?
    \item \textbf{RQ2:} Which contextual factors shape the emotional response to arguments?
\end{compactitem}
Next to answering these questions, the main contributions are the novel framework and the associated corpus \contarga,
consisting of 4000 individual annotations for 800
arguments\footnote{The data is available under a Creative Commons license here: \url{https://www.uni-bamberg.de/en/nlproc/projects/emcona/}.}.

\section{Related Work}
We now summarize relevant research on argument mining, emotion
analysis, and appraisal theories to contextualize our work.

\subsection{Argument Mining and Convincingness}
Previous work has looked in depth at the textual qualities that make
an argument convincing \citep{habernal-gurevych-2016-makes,
  habernal-gurevych-2016-argument}. However, the argument itself is
not the only driver of the argument's convincingness; both sender and
receiver play a
part. \citet{al-khatib-2020-characteristics-debaters-persuasiveness}
and \citet{lukin-etal-2017-argument} examine the personality of the
receiver using the Big Five personality traits, finding that certain
personality types are more receptive to certain types of
arguments. \citet{lukin-etal-2017-argument} also show that the prior
belief is important, with receivers who hold only weak beliefs about
the argument topic being more easily
convinced. \citet{durmus-cardie-2018-exploring} show that the prior
belief of the receiver plays a more important role for the prediction
of convincingness than linguistic features of the argument. In line
with that, \citet{rescala-etal-2024-language} demonstrate that LLMs
perform on par with humans when predicting which arguments would be
convincing for individuals with specific demographics and beliefs.

We follow this work by collecting not only assessments of argument
properties, but also properties of the annotators (age, gender,
education level and Big Five personality traits). We specifically
collect the stance of the receiver toward the argument topic
beforehand and check for stance change after an argument was
displayed. Additionally, we collect information about the sender that
annotators imagine after reading the argument (age, education level,
personality traits and a free text description).

\subsection{Emotion Analysis and Appraisals}\label{sec:rel_work_appraisals}
There are two main types of emotion models for emotion analysis:
categorical and dimensional. Ekman's basic emotion model, as an
example for categorical models, consists of six discrete emotions:
anger, surprise, disgust, joy, fear, and sadness
\citep{ekman1992-emotions}. Dimensional models represent emotions
along continuous axes in a multidimensional space. The most prominent
dimensional model in NLP is the Circumplex Model of Affect
\citep{posner2005affect}, evaluating emotions in terms of valence and
arousal.  Appraisal theories also constitute a dimensional model, in
which emotions are considered in terms of the cognitive evaluation
(appraisal) of an event \citep{scherer2001appraisal}. While multiple
frameworks of appraisal theories exist
\citep{roseman1984cognitive,roseman2001appraisal,scherer2009dynamic},
there is no consensus on the concrete set, leading to specialized
framework for comparably niche areas, such as conspiracy theories
\citep{pummerer_appraisal_2024}. Common variables in appraisal
frameworks do, however, encompass aspects of agency, pleasantness,
consequences on the self, responsibility, or expected effort and novelty.

While most work on emotion analysis is on categorical models, there is
considerable work that employs appraisal theories
\citep{klinger-2023-event}, including work on coping strategies
\citep{troiano-etal-2024-dealing}, social media analysis
\citep{stranisci-etal-2022-appreddit}, or emotion event self reports
\citep{hofmann-etal-2020-appraisal}. The largest set of appraisal
variables for emotion analysis has been proposed by
\citet{Troiano2023-crowdenvent}, which we use as inspiration in our
work. There is no appraisal framework for arguments in psychological
research (but the recognition that it would be important,
\citet{dillard_affect_2000}). Therefore, our contribution is both on
the conceptual side and the empirical natural language processing side.

\begin{table*}
    \centering
    \small\setlength{\tabcolsep}{4.5pt} 
    \begin{tabular}{lr}
    \toprule
         Dimension & Description \\
         \cmidrule(r){1-1}\cmidrule(l){2-2}
         Suddenness & the argument appears sudden or abrupt to the receiver  \\
         Suppression & the receiver tries to shut the argument out of their mind  \\
         Familiarity & the argument is familiar to the receiver  \\
         Pleasantness & the argument is pleasant for the receiver  \\
         Unpleasantness &  the argument is unpleasant for the receiver  \\
         Consequencial Importance & the argument has important consequences for the receiver  \\
         Positive Consequentiality & the argument has positive consequences for the receiver  \\
         Negative Consequentiality & the argument has negative consequences for the receiver  \\
         Consequence Manageability & the receiver can easily live with the unavoidable consequences of the argument   \\
         Internal Check & the consequences of the argument clash with the receiver's standards and ideals  \\
         External Check & the consequences of the argument violate laws or socially accepted norms  \\
         Response urgency & the receiver urges to immediately respond to the argument  \\
         Cognitive Effort & processing the argument requires a great deal of energy of the receiver  \\
         Argument Internal Check & statements in the argument clash with the receiver's standards and ideals  \\
         Argument External Check & statements in the argument violate laws or socially accepted norms  \\
         \bottomrule
    \end{tabular}
    \caption{Appraisal dimensions used to measure
    argument evaluation in the context of the self.}
    \label{tab:appraisal_descriptions}
\end{table*}

\subsection{Emotions in Arguments}
A substantial number of psychological studies point out the role of
cognitive argument evaluations for convincingness
\citep{bohner_affect-persuasion-1992,
  petty_mood_persuasion_1993,pfau_affect_resistance_2006,
  Worth1987CognitiveMO,Benlamine2015EmotionsArgumentesEmpirical}. Particularly,
\textit{pathos} plays an important role in stance changes
\citep{benlamine_2017-persuation-emotions}. In natural language
processing, research on emotions in arguments is limited, where
emotional appeal is one factor of many
\citep{habernal-gurevych-2016-makes} or is considered fallacious
\citep{evgrafova-etal-2024-analysing}.

The convincingness and emotion are related, though studies with large
language models show the challenging nature of automatically detecting
these properties \citep{chen-eger-2025-emotions}. There is very little
work that studies not only binary emotionality but differentiates
emotion categories.  \citet{greschner-klinger-2025-fearful}, as one of
the few cases, demonstrate that the discrete emotion category enhances
LLM predictions for emotionality in arguments.
\citet{habernal-gurevych-2016-makes,habernal-gurevych-2016-argument}
constructed and analyzed a corpus for convincingness strategies,
including an emotion layer. In line with that,
\citet{wachsmuth-etal-2017-computational} find a weak positive
correlation between the emotional appeal of arguments and
convincingness when analyzing 15 dimensions of arguments.
\citet{lukin-etal-2017-argument} focus on audience factors and
argument convincingness and find that information about personality
traits of the audience improves the automatic prediction of belief
change for emotional arguments.

Our work considers sender and receiver properties in the context of
argument convincingness, focusing on the implicitly evoked emotions in
receivers. To the best of our knowledge, this work is the first to
introduce appraisal theories to arguments in NLP.

\section{Contextualized Argument Appraisal Framework}
We illustrate the Contextualized Argument Appraisal Framework (\caaf)
in Figure \ref{fig:framework}. \caaf considers arguments within a
situational context with the sender and world knowledge, and the self,
the receiver. The receiver cognitively evaluates, in a synchronized
but not necessarily sequential manner, the appraisal, emotion
category, and convincingness of the argument.

\begin{figure*}
    \centering
    \includegraphics[width=\linewidth]{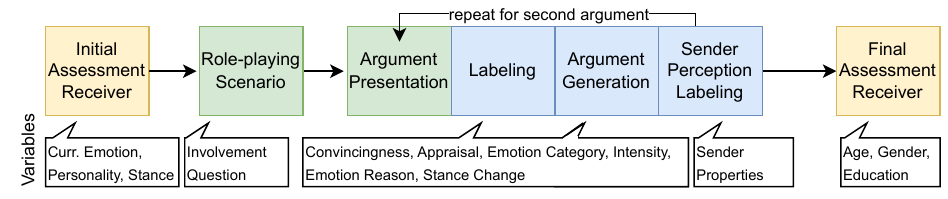}
    \caption{Flowchart of the annotation workflow. Yellow fields indicate information collected about the receiver (the annotator), green fields indicate argument and scenario display and blue fields indicate argument specific annotation.}
    \label{fig:enter-label}\label{fig:flowchart}
\end{figure*}

\subsection{Emotions}
The cognitive appraisal of the argument leads to a subjective
experience that corresponds to a discrete emotion category. This
emotion category, along with the perceived emotion intensity, are
subjective variables of the receiver and need to be assessed by them
(i.e., cannot be externally reconstructed by annotators).

Arguments are situated within context and presented by a sender. We
include these factors in \caaf to study the major source of the
emotion stimulus: \textit{how} is the emotion evoked and \textit{why}
is the emotion evoked.

\subsection{Appraisals}
To reconstruct the subjective cognitive evaluation process of a
presented argument in context, we introduce appraisal variables for
arguments, the main and novel component in our framework.
We base the appraisals used in our study on the work by
\citet{Troiano2023-crowdenvent}, adapting a subset (\suddenness,
\familiarity, \pleasantness, \unpleasantness, \intcheck, \excheck) to match the appraisal of arguments, as described in
Table~\ref{tab:appraisal_descriptions}. Furthermore, we design nine
additional appraisal dimensions: \suppression, \consimportnace, \posconseq, \negconseq,
\conseqmanage, \responseurgency, \cogeffort,
\argintcheck, and \argexcheck.

Appraisal dimensions allow us to capture the underlying cognitive
evaluations that give rise to emotional responses. They enable a
systematic study of subjective argument properties.

\subsection{Convincingness}
Central to the \caaf is the convincingness assessment of
arguments. Convincingness refers to the quality of an argument that
leads its audience to believe the content of the argument is factually
true or morally right. In our framework, we acquire this variable by a
self-assessment of the argument receiver, but, in addition, obtain the
psychologically more valid property of an observed stance change. We
do, however, use the convincingness variable in our studies due to
pragmatic design choices in the data collection described later.

\subsection{Context}
The variables above can only be assessed in the situational context,
which therefore requires information about the sender and the
receiver.  Text is rarely produced in a contextual vacuum. Rather it
is influenced by demographic properties, personality traits and
preconceived notions of the people involved. Capturing these features
allows us to create models that take individual nuances into account,
both in the ground-truth properties of the receiver (e.g., their
education level) and the perceived properties of the sender (e.g.,
does the argument sound like it was said by an educated person).

\section{Corpus Creation}
\label{corpus_creation}

We now describe our annotation setup based on \caaf, to study the
correlations of emotion, appraisal, and convincingness, and pave the
way for computational modeling. Screenshots of the annotation study are in Appendix~\ref{sec:study_screenshots_appendix}.

\subsection{Overview}
Figure~\ref{fig:flowchart} illustrates our annotation setup. The study
begins with assessing the participant's emotional state, personality,
and stance toward the topic. Subsequently, participants are immersed
in a role-playing scenario simulating a town-hall meeting to
approximate a real-world argumentative context. The exact scenario descriptions and the involvement questions are Appendix~\ref{role-playing-scenario}. 

To ensure that participants do engage with the role-playing, we ask
them to describe where they would sit and what they would see in their
surroundings. Next, we tell participants that a speaker presents an
argument, which they annotate. They then generate an argument
themselves to increase involvement (this data is not evaluated in this
paper). Participants further answer demographic and personality trait
questions about the speaker. This process is repeated for a second
argument within the same role-playing scenario. Lastly, the
participants disclose own demographics. Seven attention and sanity
checks are included in the study; see Appendix~\ref{sec:study_screenshots_appendix} for details.

\begin{table*}
    \centering
    \small
    \begin{tabularx}{\linewidth}{lXp{1.4cm}>{\centering\arraybackslash}p{1.5cm}>{\centering\arraybackslash}p{1cm}}
      \toprule
      &&& \multicolumn{2}{c}{Annotations} \\
      \cmidrule(l){4-5}
      ID & Argument & Topic & Emo. & Conv. \\
    \cmidrule(r){1-1}\cmidrule(lr){2-2}\cmidrule(lr){3-3}\cmidrule(lr){4-4}\cmidrule(l){5-5}
      1 & ``people simply can't compete with others in the same
          professional setting once they reach a certain age,
          mandatory retirement is necessary for fair competition in a
          limited job market'' & Retirement
         & \guilt\par \disgust \par \relief \par \sadness \par \surprise & 4\par1\par3\par2\par1 \\
    \cmidrule(r){1-1}\cmidrule(lr){2-2}\cmidrule(lr){3-3}\cmidrule(lr){4-4}\cmidrule(l){5-5}
      2 & ``women in polygamous marriages have been shown to be at
          much greater risk of abuse than those in monogamous
          marriages'' & Polygamy
         & \disgust\par \sadness\par\anger \par \sadness \par\surprise & 3 \par 2 \par 3 \par 3 \par 2\\
    \cmidrule(r){1-1}\cmidrule(lr){2-2}\cmidrule(lr){3-3}\cmidrule(lr){4-4}\cmidrule(l){5-5}
      3 &  ``there are direct links to drinking bottled water and
          certain types of disease, even cancer'' & Plastic bottles
         & \fear \par \trust \par \anger \par \guilt \par \surprise & 4\par 1\par1\par 2 \par 2 \\
    \bottomrule
    \end{tabularx}
    \caption{Examples from the \contarga corpus \citep[argument text
      source:][]{habernal-gurevych-2016-argument,gretz-etal-2019-IBMcorpus}. Emotion
      and convincingness assessments are from five individual
      annotators.}
    \label{tab:data_examples}
\end{table*}

\subsection{Variables}
\label{subsec:variables}
We now turn to a detailed description of the annotated
variables.

\subsubsection{Argument}
\label{subsec:argumentvariables}
The following section describes the collected variables on the
argument level in detail. All variables are assessed on a 5-point
Likert scale if not otherwise stated.

\paragraph{Emotions.}
Our emotion label set comprises Ekman's basic emotions \anger,
\disgust, \fear, \joy, \sadness, \surprise, expanded by \shame,
\guilt, \pride, 
\citep{greschner-klinger-2025-fearful}, \trust and \relief \citep{Troiano2023-crowdenvent}.  We refer to the emotion with the highest rating as
the dominant emotion. In the case of ties, participants are requested
to chose one. We additionally assess the perceived interest and boredom of
the argument, see Appendix~\ref{sec:study_screenshots_appendix} for the formulations.

\paragraph{Emotion Reason.}
Participants are instructed to provide a short textual explanation of
the evoked emotions using a free text field.

\paragraph{Emotion Influence.}
Participants answer three questions about the influence of the sender,
argument content, and receiver. The
questions are phrased in the format ``[The speaker identity, the
argument itself, who I am] was important for the emotion I
developed''.

\paragraph{Appraisals.}
We phrase each appraisal dimension as an affirmation, e.g., for
response urgency we formulate: ``I feel the urge to immediately
respond to the argument''.

\paragraph{Manipulation.}
For each argument, participants provide their perceived degree of
manipulative intent. We formulate the question: ``Do you feel
manipulated by this argument?''\footnote{In preliminary studies, we
  found perceived manipulation to decrease convincingness. In
  \contarga, there is a weak, significant negative correlation between
  manipulation and convincingness, which we leave for future work for
  a more in-depth analysis.}.

\paragraph{Convincingness.}
Similar to the manipulation variable, participants annotate the perceived
convincingness for the question ``How convincing is this argument for
you?''

\subsubsection{Receiver}\label{subsec:receiver}
Given that individual differences among receivers can affect the
perceived persuasiveness of arguments, as demonstrated by
\citet{lukin-etal-2017-argument}, our study accounts for these
variations by employing the following measurements.

\paragraph{Stance in Relation to Argument Topic.}
Participants are asked to provide their stance towards a the topic of
the displayed town hall discussion. This question is asked once prior
to argument exposure and once after each argument, allowing us to
measure potential change in the stance after specific
arguments. Additionally, participants provide their familiarity with
the topic (not to be confounded with the appraisal of familiarity with
the argument).

\paragraph{Demographics.}
Previous work on argument quality has shown that the lack of
information about the annotators of corpora poses an impediment to
perspectivist approaches to the topic
\citep{romberg-etal-2025-perspectivist-argument-quality}. To avoid
this shortcoming, we collect the age, gender and education level of
the participants.

\paragraph{Personality Traits.}
\citet{lukin-etal-2017-argument} demonstrate that more conscientious
people are rather convinced by emotional arguments. Since we
hypothesize that evoked emotions are one of the primary drivers of the
argument's convincingness, we use the Ten-Item Personality Inventory
questionnaire \citep[TIPI,][]{gosling2003very} to assess the
extraversion, agreeableness, conscientiousness, emotional stability,
and openness to experiences of the participants.

\subsubsection{Sender}\label{subsec:sender}
In our study setup, the age (old/young) and gender (male/female) of
the argument-presenting sender are systematically manipulated. While
we are aware that neither age nor gender are binary variables, this
simplified setup allows us to investigate whether these properties
influence persuasiveness. Each argument is attributed to one single
sender, whose age and gender are assigned according to a balanced
randomization scheme across topics to ensure approximately equal
representation of all four demographic combinations. We present
participants with a description of the sender and the argument in the
format ``A [speaker age] [speaker gender] approaches the microphone
and makes a statement: [argument]''.

Following the argument exposure, participants answer the same
demographic and TIPI questionnaire questions described in
Section~\ref{subsec:receiver} about their perception of the sender. This
approach allows us to examine the interplay between speaker
characteristics and audience perception within argumentative
discourse. Additionally, the participants can give details about the
imagined sender in a free text field. We leave a more in-depth
analysis of these variables for future work.

\begin{figure}
    \centering
    \includegraphics[width=\linewidth]{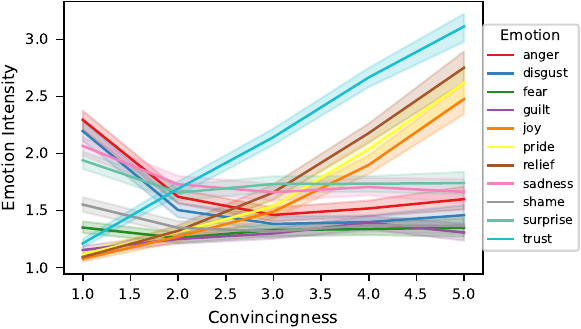}
    \caption{Emotion Intensities and Convincingness.}
    \label{fig:emotion_intensities}
\end{figure}

\subsection{Crowd-sourcing Details}\label{sec:crowd-sourcing}
We use the platform Prolific\footnote{\url{https://www.prolific.com/}}
with a self-implemented study framework based on
Streamlit\footnote{\url{https://docs.streamlit.io/}}. Participants are
required to live in the United Kingdom or Ireland, have English as
their first, native, and primary language, and have an approval rate
of 95--100\%.  Each participant answers the survey for two
arguments. We pay each participant 3.48~\pounds~for one survey, which
on average takes 22.5 minutes. The study includes 7 attention
checks. Participants can participate up to 40 times, and therefore
annotate up to 80 arguments. In total, the cost of the study amounts
to 9,404~\pounds~. The contributing participants in our studies were
on average 40.4 years old (18 minimum, 81 maximum). From this set,
1,715 identified as female, 1,541 as male, 374 as non-binary, and 30
as genderfluid/non-conforming. Further, 14 specified their own gender,
12 preferred not to answer the question, and 8 as
questioning\footnote{Note that on the Prolific platform, participants
  cannot provide their gender in such detail. We distributed our study
  as 40\% female, 40\% male, 20\% non-binary and obtained the gender
  variable in the study.}.

\subsection{Datasets}
The \textit{UKPConvArgv1} \citep{habernal-gurevych-2016-argument} and \textit{IBM-Rank-30k} \citep{gretz-etal-2019-IBMcorpus} corpora, hereafter UKP and IBM, serve as sources for the arguments. Both datasets provide isolated arguments and their respective discussion topic. Isolated arguments allow us to investigate the initial argument appraisal without adding noise due to lengthy debates containing multiple arguments. We sample 14 and 26 topics from UKP and IBM, respectively. For each topic, an equal number of pro and con arguments is sampled. 

Arguments are manually annotated for comprehensiveness, persuasive attempt, and stance alignment. Details on the manual filtering are in Appendix~\ref{appendix:manual_data_fitlering}. Importantly, we exclude arguments that point toward the age or gender of the sender, since we manipulate these variables in our study (see Section~\ref{subsec:sender}). We randomly select 800 arguments from the 953 valid ones for our study.

\begin{figure}
    \centering
    \includegraphics[width=1\linewidth]{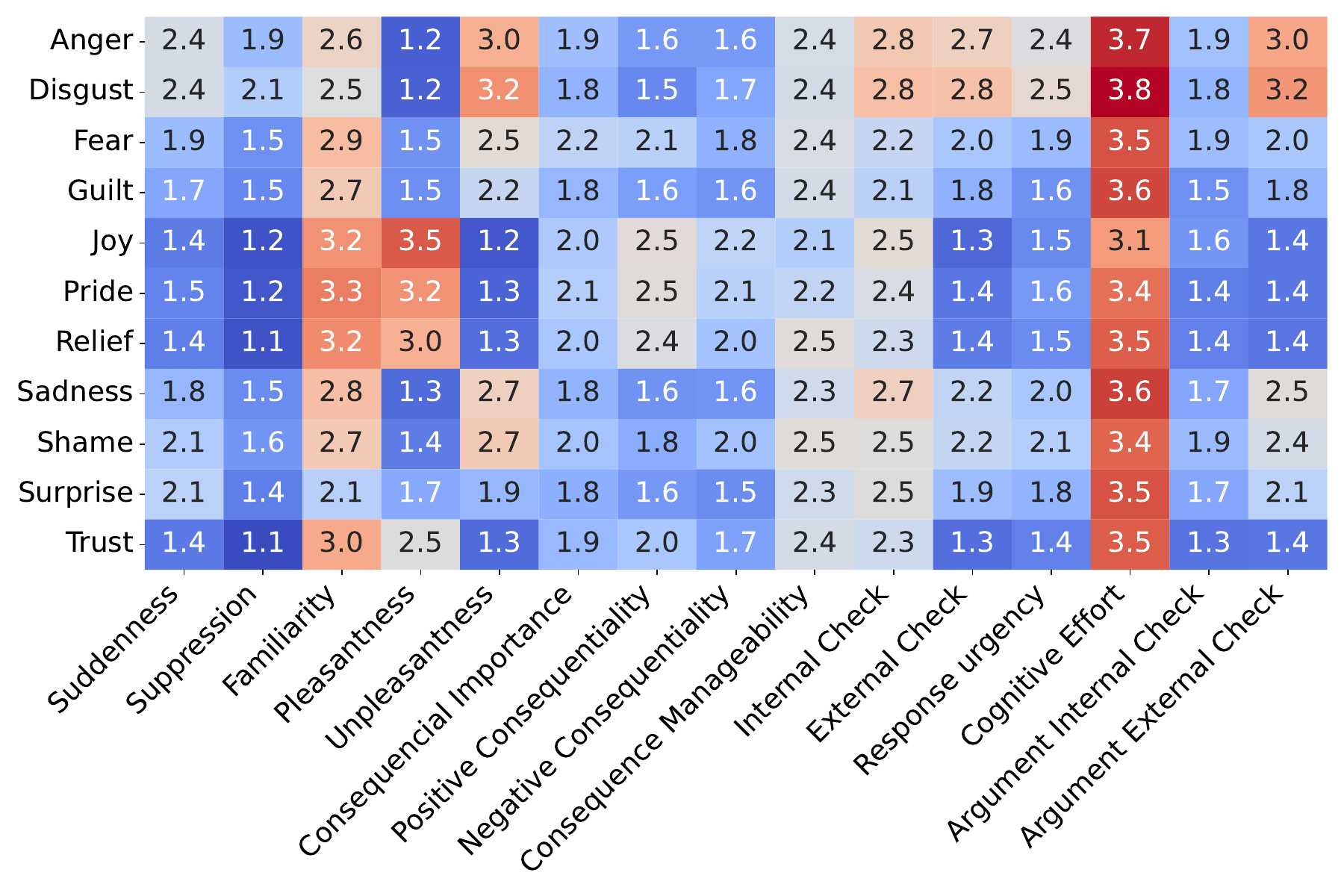}
    \caption{Analysis of discrete emotion categories and appraisal dimensions in arguments. Each row is an appraisal, each column is an emotion, and each cell is the average appraisal value for that emotion.}
    \label{fig:appraisal_heatmap}
\end{figure}

\section{Data Analysis}
Our study results in a corpus of 4,000 individual argument annotations
of 800 arguments from 39 topics with 5 individual annotations (170
annotations excluded due to failed attention checks.
Details about the topics can be found in
Appendix~\ref{appendix:topics}.

Table~\ref{tab:data_examples} displays example arguments with emotion
and convincingness labels. The average argument length is
32.1 tokens, 1.7 sentences. The average value of convincingness is 2.7
(min=1, max=5); the values vary across topics, with
\textit{retirement} being most (3.2) and \textit{atheism} being least
convincing (2.0).  The frequencies of the dominant emotions are, in
descending order: \trust (709), \surprise (564), \sadness (561),
\anger (521), \relief (382), \disgust (292), \pride(198), \joy (193),
\shame (100), \fear (90), \guilt (84). The average emotion intensity
value across emotions is 1.61, with \trust showing the highest (2.02), and
\guilt the lowest (1.27) value.

We now turn to the example arguments from \contarga, displayed in
Table~\ref{tab:data_examples}, in more detail. The first argument
evokes 5 different emotions and convincingness labels from
1--4. In contrast, the second argument evokes negative emotions
(\disgust, \sadness, \anger) with similar convincingness values, with the
exception of \surprise. The second argument discusses the greater risk of abuse of women in polygamous marriages, intuitively evoking negative emotions. \Surprise may arise if the receiver is unfamiliar with the topic. 

For the third argument, we find negative emotions (\fear, \anger,
\guilt). \Fear may result from the risk of developing
cancer, while a person regularly using plastic bottles could feel
guilty. The argument evoking trust is less intuitively
clear. These examples illustrate the wide variety of emotions evoked in different participants by the same arguments.

\paragraph{Inter-Annotator Agreement.}
The contextualized argument appraisal framework is designed to capture
individual argument appraisals on divisive topics. We intentionally
construct it to elicit differences between receivers of arguments
rather than to derive one `gold-label', leading to 5 individual argument appraisals for each argument. As expected, the
inter-annotator agreement on single variables is therefore low, with a
Krippendorff's Alpha of 0.06 for convincingness and Fleiss' Kappa of
0.17 for the most dominant emotion category of an argument. These
numbers do not suggest a low quality of the data. In contrast, they justify the need
for modeling and analysis under the paradigm of perspectivism.

\begin{table}[t]
  \centering\small\setlength{\tabcolsep}{3pt}
  \newcommand{\p}{\phantom{$-$}}
    \begin{tabular}{llll}
    \toprule
    Emotions & \multicolumn{1}{c}{$r$} & Appraisals & \multicolumn{1}{c}{$r$} \\
    \cmidrule(r){1-1}\cmidrule(lr){2-2}\cmidrule(lr){3-3}\cmidrule(l){4-4}
    Trust & \p.58*** & Pleas & \p.58*** \\
    Relief & \p.51*** & Pos Cons & \p.40*** \\
    Pride & \p.47*** & Fam & \p.34*** \\
    Joy & \p.45*** & Neg Cons & \p.22*** \\
    Guilt & \p.11*** & Cons Imp & \p.15*** \\
    Fear & \p.01  & Cons Man & $-$.02   \\
    Shame & $-$.05* & Cog Eff & $-$.08*** \\
    Surprise & $-$.09*** & Int Ch & $-$.11*** \\
    Sadness & $-$.13*** & Arg Int Ch & $-$.12*** \\
    Anger & $-$.22*** & Resp Ur & $-$.25*** \\
    Disgust & $-$.23*** & Sup & $-$.30*** \\
    \cmidrule(r){1-1}\cmidrule(l){2-2}
      Other & \multicolumn{1}{c}{$r$} & Sud & $-$.34*** \\
    \cmidrule(r){1-1}\cmidrule(l){2-2}
    Interest & \p.69*** & Unpleas & $-$.36*** \\
    Manipulation & $-$.10*** & Ex Ch & $-$.37*** \\
    Boredom & $-$.49*** & Arg Ex Ch & $-$.50*** \\
    \bottomrule
    \end{tabular}
    \caption{Pearson correlations ($r$) of emotions and appraisals
      with the convincingness variable. Significance levels (with
      Bonferroni correction): * $p < .05$, ** $p < .01$, ***
      $p < .001$.}
    \label{tab:correlations_table}
\end{table}

\section{Results}
In the following, we answer our posed research questions individually.

\subsection{RQ1: How do discrete emotions and appraisals correlate with argument convincingness?}\label{rq1}
We analyze \contarga to investigate the interplay between evoked emotions and convincingness.

\paragraph{Emotions.}
We aim at understanding the specific emotions that are evoked in the
receivers of arguments to examine their interplay with perceived
convincingness. To this end, we analyze \contarga arguments for relationships between
discrete emotion categories and convincingness labels.

Table~\ref{tab:correlations_table} displays all correlation values of the variables of interest and convincingness with Bonferroni correction. All correlations between emotion and convincingness are significant, except for \fear. The strongest positive effects are with \trust (.57), \relief (.51), \pride (.46), and \joy
(.44). In contrast, \anger ($-$.22) and \disgust ($-$.23) show the strongest
negative correlations with convincingness.

\paragraph{Emotion Intensities.}
Figure~\ref{fig:emotion_intensities} allows a closer look at these
correlations, based on intensity values. For \trust, the emotion
intensity and convincingness have an almost linear relationship,
similarly for \relief, \pride, \joy. For \fear, \shame, and \guilt,
the emotion intensity stays relatively stable across convincingness
values. In contrast, for \anger and \disgust, the highest intensity
values are found for the lowest convincingness values, with a drop at value 2 (little convincing), continuing in a
moderate intensity (around 1.4) across higher convincingness
values. Similar, yet weaker trends are found for
\sadness and \surprise, with on average higher emotion intensity
values (around 1.7).

\begin{table}
\centering\small
\begin{tabular}{lrrr}
\toprule
Emotion & Sender & Argument & Receiver \\
\cmidrule(r){1-1}\cmidrule(lr){2-2}\cmidrule(lr){3-3}\cmidrule(l){4-4}
Anger & 1.86 & 3.78 & 3.23 \\
Disgust & 1.82 & 3.72 & 3.25 \\
Fear & 1.78 & 3.68 & 3.11 \\
Guilt & 1.65 & 3.27 & 2.81 \\
Joy & 2.13 & 3.73 & 3.21 \\
Pride & 2.24 & 3.79 & 3.24 \\
Relief & 1.98 & 3.70 & 3.14 \\
Sadness & 1.84 & 3.41 & 3.09 \\
Shame & 2.00 & 3.34 & 2.86 \\
Surprise & 1.86 & 3.39 & 2.69 \\
Trust & 1.82 & 3.47 & 2.97 \\
\cmidrule(r){1-1}\cmidrule(lr){2-2}\cmidrule(lr){3-3}\cmidrule(l){4-4}
Avg. & 1.89 & 3.56 & 3.05 \\
\bottomrule
\end{tabular}
\caption{Average values of how important the sender, the argument, and
  the receiver are for the emotional response of the receiver. The
  importance of each dimension is rated on a 1--5
  scale.}\label{tab:emotion_influence}
\end{table}

\paragraph{Appraisals.}
We now focus on the analysis of emotions and appraisals in arguments,
illustrated in Figure~\ref{fig:appraisal_heatmap}. For each emotion,
the average appraisal value is displayed in the heatmap. To the best of our knowledge, we are the first to apply appraisal theory to arguments. However, the appraisal dimensions of \suddenness, \familiarity, \pleasantness, \unpleasantness, \intcheck, \excheck values
are in line with other applications of appraisal theories to emotion
analysis in natural language processing \citep[Figure 8]{Troiano2023-crowdenvent}, confirming the validity of the approach. Novel in our approach is the inclusion of more appraisal dimensions and the shift from evaluating events to evaluating arguments.

Turning to the analysis of appraisal dimensions in arguments, we find that \unpleasantness, \intcheck and \excheck appraisals consistently show
the highest ratings for \anger (3.0, 2.8, 2.7) and \disgust (3.2, 2.8,
2.9). Intuitively, arguments evoking negative emotions are being
appraised as unpleasant, clashing with the receiver's ideals, and
violating social norms. In contrast, \joy, \pride, and \relief show
high values for \pleasantness (3.5, 3.2, 3.0), \familiarity (3.1, 3.3,
3.2), and \posconseq (2.5, 2.5, 2.5). This indicates that arguments being appraised as beneficial, familiar, and rewarding evoke positive
emotions.

Beyond categorical emotions, appraisals are also related with convincingness. Table~\ref{tab:correlations_table} displays all correlations. We find significant correlations for all appraisal dimensions except
\conseqmanage. \Pleasantness shows a strong positive (.57), \posconseq a
moderate (.39), and \familiarity a moderate (.33) positive correlation
with convincingness. In contrast, we find strong negative correlations
with \argexcheck($-$.50), \unpleasantness ($-$.39), as well as \excheck
($-$.36).

Both discrete emotion and appraisal analyses demonstrate significant
correlations between positive emotions and convincingness, and
negative correlations between negative emotions and
convincingness. The results with respect to discrete emotions align with
\citet{greschner-klinger-2025-fearful} (showing negative correlations between negative emotions and convincingness using 300 German arguments), but contradict that \fear
increases convincingness \citep{dillard2004role}.

\subsection{RQ2: Which contextual factors shape the emotional response
  to arguments?}\label{rq2}
\label{rq2}
After analyzing \textit{which} emotions are developed and how they are
correlate with the argument's convincingness, we now turn to
understanding \textit{how} these emotions are developed. We analyze the participant's (argument receiver's) textual explanations about the evoked emotions and the
strongest influence on the emotion (sender, argument, receiver). We also
analyze how strong (1--5) the influence (sender, argument,
receiver) was for the development of the evoked emotion.

Table~\ref{tab:emotion_influence} shows the results. The argument
(3.6) is the primary driver of the emotional response, followed by the
receiver themselves (3.1) and the sender (1.9). Receivers rarely
attribute \guilt and \shame to being evoked by the sender. \Pride is often attributed to the argument itself, which is also the
most frequent emotion and has the highest correlation to
convincingness. Similar patterns are found for \anger, which has,
however, the strongest negative convincingness correlation. Interestingly, the sender is the least important component when it comes to the evoked emotion of \guilt. In
summary, the argument itself is the primary driver of the emotional
response.

\section{Conclusion}
In this paper, we introduce the Contextualized Argument Appraisal
Framework (\caaf). It is a novel approach to explain how arguments are
cognitively appraised. In a synchronized process, individuals develop
an emotion and perceive a particular convincingness after argument
exposure. The analysis of the resulting corpus \contarga, the data
resulting from a role-playing annotation study based on \caaf, shows
that appraisals are appropriate for explaining the relationship
between emotions and convincingness. Furthermore, our results validate
the framework's contextual approach: while the arguments themselves
generally exert the strongest influence on emotional responses,
emotion-specific patterns emerge regarding the relative contributions
of sender, argument, and receiver factors. We release the data,
enabling research on arguments with focus on such contextual
properties.

Our work demonstrates that affective responses are not merely
superficial reactions but are integrally tied to how arguments are
processed. Future work shall study \caaf from a
computational perspective. We hypothesize that the integration of
appraisal and categorical emotion variables improves the performance of
automatic convincingness estimates and make the decision process
transparent. Further, models that automatically assess the subjective
appraisal and emotion are valuable in itself for corpus and
argumentation studies.

Particularly, \caaf helps to disentangle the influences that shape a
person's reaction to arguments, because it captures the role of
properties of the sender, receiver, and the argument itself. In
downstream systems, these factors can contribute to argument mining
and content moderation systems, for example to support studies on
desinformation and argumentation strategies. The framework lays the
groundwork for context-sensitive and emotionally aware computational
systems that reflect how people engage with persuasive text.

\section*{Limitations}
One of the limitations of this work is the focus on the English
language and a relatively limited geographical spread of the
annotators involved in the creation of our corpus. We make this choice
to ensure higher data quality, limiting a possible mismatch between
dialects or regional language usage. On the other hand, a wider
linguistic and geographic spread would allow us to make statements
about the nature of emotions in arguments that are more
comprehensive. While we see these shortcomings in the annotation
itself, the Contextualized Argument Appraisal Framework stands
independent of its implementation in this specific annotation scenario
and can be used across various languages.

Previous work has also examined multi-turn argumentative
conversations. While this allows an observation of how emotions and
appraisals develop over the course of a longer exchange, this is out
of the scope of our current setting of the framework. Adapting it for
multi-turn interactions is another avenue for future work.

Regarding our study, each participant annotates two arguments from one
topic. The argument selection from the topic is randomized, which can
lead to biases due to the prior stance of the participant and the
order of pro and con arguments. We acknowledge that future work can
take a deep dive into the role of argument presentation according to
stances, since we report the exact argument in displayed order within
our available \contarga corpus.  Further, we allow participation up to
39 times, which ensures that one participant can only annotate two
arguments per topic, however, there might be annotators that do in
fact participate in all 39 studies (i.e, contributing 78 individual
argument annotations out of the 4,000 arguments in \contarga), therefore, introducing annotator bias.

To address the emerging challenge of automated bot participation in crowdsourcing studies, we manually verfy all free-text response to ensure authenticity. Participants whose completion times fell significantly below the study average were systematically excluded from analysis.
Recognizing the inherent quality control challenges in crowdsourced research, we incorporated seven attention and saliency checks throughout the study. Only data from participants who successfully passed all validation measures were retained for analysis. These quality assurance protocols, while necessary to mitigate common crowdsourcing limitations, enabled us to leverage the key advantage of crowdsourcing: access to a demographically diverse participant pool that would be difficult to achieve through other recruitment methods.

\section*{Ethical Considerations}
The study is part of a larger project that has been approved,
including a description of the presented study, by the ethics board of
the University of Stuttgart. Following recommendations with respect
to ethical challenges in emotion analysis
\citep{mohammad-2022-ethics-sheet}, we point out the following facts
about our work: The purpose of the \caaf is to study the interplay of
emotions in argument texts while taking contextual factors into
account, focusing on the emotions evoked in receivers. We choose
categorical emotion labels from emotion theories in accordance with
previous work, and expand appraisal labels from dimensional appraisal
theories to be applicable to arguments.

We employed crowd-sourcing to collect human annotations for the
cognitive appraisal process of arguments. For a given argument,
participants provide emotional responses, convincingness, stances
toward and familiarity with the topic. In addition, participants
provide demographic information, personality traits, and their
emotional state before starting the study. We informed the
participants that the intent of the study is research with the goal of
a scientific publication. We obtained consent from all participants to
use their data to analyze emotions and convincingness in arguments as
well as publishing their annotation in anonymized from. While there is
one free text field in our study, we manually check all of these
responses for any hints that would allow tracing back the identity of
the annotator, for instance, if a participant unconsciously disclosed
personal information.  We explicitly make annotators aware that they
will not be compensated if they fail more than 2 attention
checks. Details about the payment are described in
Section~\ref{sec:crowd-sourcing}. We are aware that being exposed to
arguments can be upsetting. We manually select a comprehensive set of
trigger words and warn the study participants about the potentially
upsetting nature of arguments. The participants can drop out of the
study at any point without consequences.

For this study, we chose to use crowd-sourcing since the study is closer to a study in psychology than NLP. Our data collection can be seen more as an online experiment. The platform we use, Prolific, is designed for that. However, we do provide fair compensation to our study participants and report all relevant details about the online experiment. The quality of our data is ensured by multiple attention and saliency checks, which are all reported. With proper attention and saliency checks as well as a careful study design, crowdsourcing is the best option to collect lots of data from a broad variety of people with different demographics.

\section{Acknowledgements}
This project has been conducted as part of the \textsc{Emcona} (The Interplay of Emotions and Convincingness in Arguments) project, which is funded by the German Research Foundation (DFG, project KL2869/12--1, project number 516512112). We thank Meike Bauer for her contribution to the data preparation and the fruitful discussions.

\section{Bibliographical References}
\label{sec:reference}

\bibliographystyle{lrec2026-natbib}
\bibliography{lit}

\bibliographystylelanguageresource{lrec2026-natbib}
\bibliographylanguageresource{languageresource}

\appendix

\section{Manual Data Filtering of UKP and IBM.}
\label{appendix:manual_data_fitlering}
The \textit{UKPConvArgv1} \citep{habernal-gurevych-2016-argument} and
\textit{IBM-Rank-30k} \citep{gretz-etal-2019-IBMcorpus} corpora,
further referred to as UKP and IBM, serve as the source for the
arguments. We aimed for 1000 arguments in total. We approximate
needing 40 various topics to capture topical differences; hence, we
select all 16 topics from UKP and randomly select 24 topics from
IBM. For each topic, we randomly sample 14 arguments per stance
(pro/con). For this set, we perform standard data cleaning and
pre-processing steps.

We manually check if the argument matches our definition of arguments: \textit{The argument must provide reasons supporting a specific claim (standpoint or point of view). The argument must be complete and comprehensive in itself. The reasons must at least be attempted to be logical and reasonable. The argument must follow the goal of persuasion (convince others of the supported standpoint) and must be logical in itself.}
Further, we exclude arguments using the following criteria. (1) The argument is not in line with our definition of an argument. (2) The argument shows hints toward author characteristics (e.g., ``As a woman, I would argue that...''). (3) The argument is a duplicate. (4) The argument does not have a clear stance toward the topic. 

For the resulting arguments, we check the stance of each argument for correctness, re-annotate the stance if necessary, and normalize all stances to be either “yes” if the argument supports the topic or “no” if the argument opposes the topic. We normalize the arguments’ topics according to this structure: ``We should [verb] [object].'' For instance, we normalize the topic “gay-marriage-right-or-wrong” to “We should allow gay marriage”. 

This manual effort results in 953 arguments, from which we randomly select 800 for the annotation.

\section{Correlations}\label{appendix:all_correlations}
To answer our research questions, we compute correlations between all emotion variables, interest, boredom, manipulation, all appraisal variables, and convincingness. The results are displayed in Table~\ref{tab:appendix:correlations_table}. The correlations are contextualized, interpreted, and analyzed in the main part of the paper.

\begin{table}[t]
\centering\small\setlength{\tabcolsep}{3pt} 
\newcommand{\pstar}{\phantom{***}}
\newcommand{\ptwostars}{\phantom{**}}
\begin{tabular}{lrrrrr}
\toprule
Variable & \multicolumn{1}{c}{$r$} & \multicolumn{1}{c}{$p$}  \\
\cmidrule(r){1-1}\cmidrule(lr){2-2}\cmidrule(l){3-3}
Anger & $-$0.222***  & 0.000   \\
Disgust & $-$0.233***  & 0.000   \\
Fear & 0.006\pstar  & 1.000  \\
Guilt & 0.113***  & 0.000   \\
Joy & 0.447***  & 0.000   \\
Pride & 0.465***  & 0.000   \\
Relief & 0.510***  & 0.000   \\
Sadness & $-$0.125***  & 0.000   \\
Shame & $-$0.052*\ptwostars  & 0.028   \\
Surprise & $-$0.085***  & 0.000   \\
Trust & 0.578***  & 0.000   \\
Suddenness & $-$0.345***  & 0.000   \\
Suppression & $-$0.296***  & 0.000   \\
Familiarity & 0.336***  & 0.000   \\
Pleasantness & 0.576*** & 0.000   \\
Unpleasantness & $-$0.363***  & 0.000   \\
Consequential Importance & 0.151***  & 0.000   \\
Positive Consequentiality & 0.400***  & 0.000   \\
Negative Consequentiality & 0.217***  & 0.000   \\
Consequence Manageability & $-$0.018\pstar  & 1.000   \\
Internal Check & $-$0.105***  & 0.000   \\
External Check & $-$0.366***  & 0.000   \\
Response Urgency & $-$0.249***  & 0.000   \\
Cognitive Effort & $-$0.078***  & 0.000   \\
Argument Internal Check & $-$0.124***  & 0.000   \\
Argument External Check & $-$0.498*** & 0.000   \\
Manipulation & $-$0.104***  & 0.000   \\
Interest & 0.691*** & 0.000   \\
Boredom & $-$0.492*** & 0.000   \\
\hline
\end{tabular}
\caption{Pearson correlations ($r$) and corresponding $p$ values with Bonferroni correction
      for all variables discussed in the main paper. Significance levels: * $p < .05$, ** $p < .01$, ***
      $p < .001$. All numbers are rounded to three decimal points.}
\label{tab:appendix:correlations_table}
\end{table}

\section{Emotion Influence}\label{appendix:emotion_influence}
In Section~\ref{rq2} we analyze the importance of the sender,
argument, and receiver in detail to understand the drivers of
emotional responses. We analyze how important (1-5) the influence
(sender, argument, receiver) was for the development of the evoked
emotion, individually for each topic and display the results in
Table~\ref{appendixtab:topic_emotion_influence}. While the main
results are discussed and contextualized in Section~\ref{rq2}, we
display the results for all topics here.
\begin{table}[t]
\centering\small\setlength{\tabcolsep}{3pt} 
\begin{tabular}{lrrr}
\toprule
Topic & Sender & Argument & Receiver \\
\cmidrule(r){1-1}\cmidrule(lr){2-2}\cmidrule(lr){3-3}\cmidrule(l){4-4}
Weapons & 1.67 & 4.08 & 3.30 \\
Farming & 1.48 & 3.76 & 3.37 \\
Zoos & 1.51 & 3.71 & 3.05 \\
Libertarianism & 1.54 & 3.69 & 3.18 \\
Gay Marriage & 1.85 & 3.97 & 3.71 \\
Guantanamo & 1.65 & 3.76 & 3.16 \\
Cosmetic Surgery & 1.78 & 3.73 & 2.89 \\
Exec. Compensation & 1.64 & 3.51 & 3.02 \\
Gender & 2.04 & 3.87 & 3.30 \\
Naturopathy & 1.73 & 3.56 & 2.93 \\
Plastic Bottles & 1.82 & 3.63 & 2.67 \\
TV Books & 1.66 & 3.47 & 2.62 \\
Int. Property & 1.69 & 3.48 & 2.66 \\
Entrapment & 1.79 & 3.56 & 3.34 \\
Personal Pursuit & 1.80 & 3.55 & 2.93 \\
ZTP Schools & 1.76 & 3.46 & 3.04 \\
PE Schools & 1.74 & 3.43 & 3.12 \\
Atheism & 1.78 & 3.47 & 3.00 \\
Journalism & 1.76 & 3.43 & 2.81 \\
Holocaust & 1.82 & 3.48 & 2.96 \\
Economic Sanctions & 1.82 & 3.46 & 3.06 \\
Prostitution & 2.09 & 3.72 & 3.24 \\
Spouse & 2.03 & 3.66 & 3.45 \\
Creationism & 1.56 & 3.17 & 3.10 \\
Education & 1.98 & 3.57 & 3.07 \\
Spanking & 2.04 & 3.61 & 3.33 \\
Polygamy & 1.90 & 3.44 & 2.98 \\
India & 1.92 & 3.44 & 2.98 \\
School Uniform & 2.00 & 3.51 & 2.78 \\
Abortion & 2.35 & 3.85 & 3.00 \\
Women Combat & 2.31 & 3.77 & 3.39 \\
Student Loans & 1.93 & 3.36 & 2.82 \\
Judicial Activism & 2.08 & 3.50 & 2.83 \\
Fastfood & 1.95 & 3.36 & 3.09 \\
Father & 2.23 & 3.44 & 3.20 \\
Retirement & 2.38 & 3.56 & 2.84 \\
Subsidize Dads & 2.18 & 3.35 & 2.98 \\
Porn & 1.92 & 2.83 & 3.08 \\
Firefox & 2.22 & 3.14 & 2.56 \\
\bottomrule
\end{tabular}
\caption{Average emotion influence scores across topics.}\label{appendixtab:topic_emotion_influence}
\end{table}

  \begin{figure}[]
    \centering
    \includegraphics[width=\linewidth]{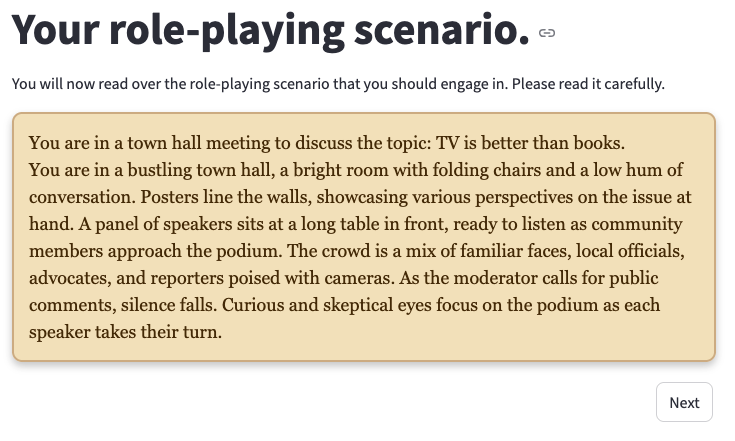}
    \caption{Description of the role-playing scenario in our annotation study.}\label{role-playing_screenshot}
  \end{figure}

  \begin{figure}[]
    \centering
    \includegraphics[width=\linewidth]{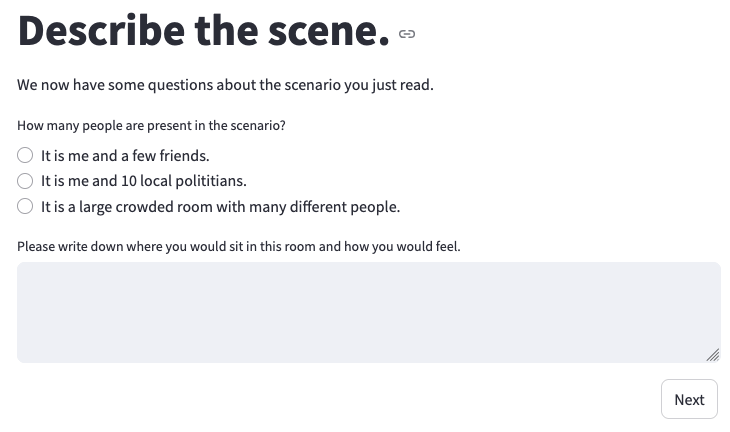}
    \caption{Involement question for the role-playing scenario in our annotation study.}\label{role-playing_engagement_screenshot}
  \end{figure}

\section{Topics}\label{appendix:topics}
\contarga contains 39 distinct topics. We display the short names for the topics, along with a description description of each topic, the number of pro, con, and total arguments in Table~\ref{tab:appendix_table_topics}.
\begin{table*}[]
    \centering\small
    \begin{tabularx}{\linewidth}{l Xrrrr}
    \toprule
    Name & Description & Corpus & Pro & Con & Total \\
    \cmidrule(r){1-1}\cmidrule(lr){2-2}\cmidrule(lr){3-3}\cmidrule(lr){4-4}\cmidrule(lr){5-5}\cmidrule(l){6-6}
    Abortion & We should legalize abortion & IBM & 9 & 9 & 18 \\
    Atheism & We should adopt atheism & UKP & 7& 7& 14\\
    Cosmetic Surgery & We should ban cosmetic surgery for minors &IBM & 7& 7&14 \\
    Creationism & We should believe in creationism & UKP& 13& 13& 26\\
    Economic Sanctions & We should end the use of economic sanctions & IBM& 11& 11& 22\\
    Education & We should subsidize vocational education & IBM & 13& 13& 26\\
    Entrapment & Entrapment should be legalized &IBM & 13& 13& 26\\
    Exec. Compensation & We should limit executive compensation &IBM & 13& 13& 26\\
    Farming & We should ban factory farming & IBM &11 &11 & 22\\
    Fastfood & We should ban fast food & IBM & 14 & 14& 28\\
    Father & It is better to have a lousy father than to be fatherless & UKP& 13& 13& 26\\
    Firefox & We should choose firefox over internet explorer &UKP & 11& 11& 22\\
    Gay Marriage & We should allow gay marriage &UKP & 4& 4& 8\\
    Gender & We should adopt gender-neutral language & IBM& 12& 12& 24\\
    Guantanamo & We should close Guantanamo Bay detention camp & IBM& 14&14 &28 \\
    Holocaust & Holocaust denial should be a criminal offence & IBM& 13& 13& 26\\
    India & India has the potential to lead the world &UKP &6 &6 &12 \\
    Int. Property & We should abolish intellectual property rights & IBM& 12& 12& 24\\
    Journalism & We should subsidize journalism & IBM& 12& 12& 24\\
    Judicial Activism & We should limit judicial activism & IBM& 13& 13& 26\\
    Libertarianism & We should adopt libertarianism &IBM & 12& 12& 14\\
    Naturopathy & We should ban naturopathy &IBM & 14& 14&28 \\
    PE Schools & Physical education should be mandatory in schools & UKP& 9& 9& 18\\
    Personal Pursuit & We should prioritize the personal pursuit over advancing the common good &UKP & 13& 13& 26\\
    Plastic Bottles & We should ban plastic water bottles &UKP & 6& 6& 12\\
    Polygamy & We should legalize polygamy & IBM &13 & 13 & 26 \\
    Porn & Porn is wrong &UKP & 8 &8 &16 \\
    Prostitution & We should legalize prostitution & IBM& 14& 14& 28\\
    Retirement & We should end mandatory retirement & IBM& 13& 13& 26\\
    School Uniform & The school uniform is a good idea & UKP& 9& 9& 18\\
    Spanking & We should allow spanking of children & UKP& 10& 10& 20\\
    Spouse & One should turn in their spouse if they committed murder & UKP& 12& 12& 24\\
    Student Loans & We should subsidize student loans &IBM & 13& 13& 26\\
    Subsidize Dads & We should subsidize stay-at-home dads & IBM &13 & 13& 26 \\
    TV Books & TV is better than books &UKP &7 &7 &14 \\
    Weapons & We should fight for the abolition of nuclear weapons &IBM & 11& 11& 22\\
    Women Combat & We should prohibit women in combat &IBM & 12& 12& 24\\
    ZTP Schools & We should adopt a zero-tolerance policy in schools &IBM & 13 &13& 26\\
    Zoos & We should abolish zoos & IBM & 13& 13& 26\\
    \bottomrule
    \end{tabularx}
    \caption{List of topics in \contarga, along with original data source (IBM-Rank-30k \citep{gretz-etal-2019-IBMcorpus} or UKPConvArgv1 \citep{habernal-gurevych-2016-argument}), pro arguments, con arguments, and total number of arguments.}
    \label{tab:appendix_table_topics}
\end{table*}

\section{Role-Playing Scenario}\label{role-playing-scenario}
To mimic the exposure to an argument in the wild and to increase emotional engagement, we create the role-playing scenario of a town hall meeting that the annotator is part of.
The exact role-playing scenario is described in Figure~\ref{role-playing_screenshot}. The involvement questions that help the annotator engage with the argument can be seen in Figure~\ref{role-playing_engagement_screenshot}.


\newpage

\section{Study Design}\label{sec:study_screenshots_appendix}
The following section shows the screenshots of out study as described in Section~\ref{sec:crowd-sourcing}. After Figure 26 the study loops back to Figure 14 to display the second argument. 

\raggedbottom

\foreach \n in {1,...,29} {
  \begin{figure}[]
    \centering
    \includegraphics[width=\linewidth]{\n.png}
    \caption{Screenshot of our study, Page No.\ \n}
  \end{figure}
}

\end{document}